\documentclass{bmvc2k}

\title{PerSense: Training-Free Personalized Instance Segmentation in Dense Images}

\addauthor{Muhammad Ibraheem Siddiqui}{muhammad.siddiqui@mbzuai.ac.ae}{1}
\addauthor{Muhammad Umer Sheikh}{muhammad.sheikh@mbzuai.ac.ae}{1}
\addauthor{Hassan Abid}{hassan.abid@mbzuai.ac.ae}{1}
\addauthor{Muhammad Haris Khan}{muhammad.haris@mbzuai.ac.ae}{1}

\addinstitution{
Mohamed Bin Zayed University of\\
 Artificial Intelligence\\
 Abu Dhabi, UAE
}
\runninghead{Siddiqui et.al}{Personalized instance Segmentation in dense images}

\usepackage[linesnumbered,ruled,vlined]{algorithm2e}
\RequirePackage{multirow}
\RequirePackage{amssymb}
\RequirePackage{booktabs}

\begin{document}

\maketitle

\begin{abstract}
The emergence of foundational models has significantly advanced segmentation approaches. However, challenges still remain in dense scenarios, where occlusions, scale variations, and clutter impede precise instance delineation. To address this, we propose \textbf{PerSense}, an end-to-end, training-free, and model-agnostic one-shot framework for \textbf{Per}sonalized instance \textbf{S}egmentation in d\textbf{ense} images. We start with developing a new baseline capable of automatically generating instance-level point prompts via proposing a novel Instance Detection Module (IDM) that leverages density maps (DMs), encapsulating spatial distribution of objects in an image. To reduce false positives, we design the Point Prompt Selection Module (PPSM), which refines the output of IDM based on adaptive threshold and spatial gating. Both IDM and PPSM seamlessly integrate into our model-agnostic framework. Furthermore, we introduce a feedback mechanism that enables PerSense to improve the accuracy of DMs by automating the exemplar selection process for DM generation. Finally, to advance research in this relatively underexplored area, we introduce PerSense-D, an evaluation benchmark for instance segmentation in dense images. Our extensive experiments establish PerSense's superiority over SOTA in dense settings. Code is available at \href{https://github.com/Muhammad-Ibraheem-Siddiqui/PerSense}{GitHub}.
\end{abstract}

%-------------------------------------------------------------------------
\begin{figure}[ht]
    \centering
    \includegraphics[width=0.97\linewidth]{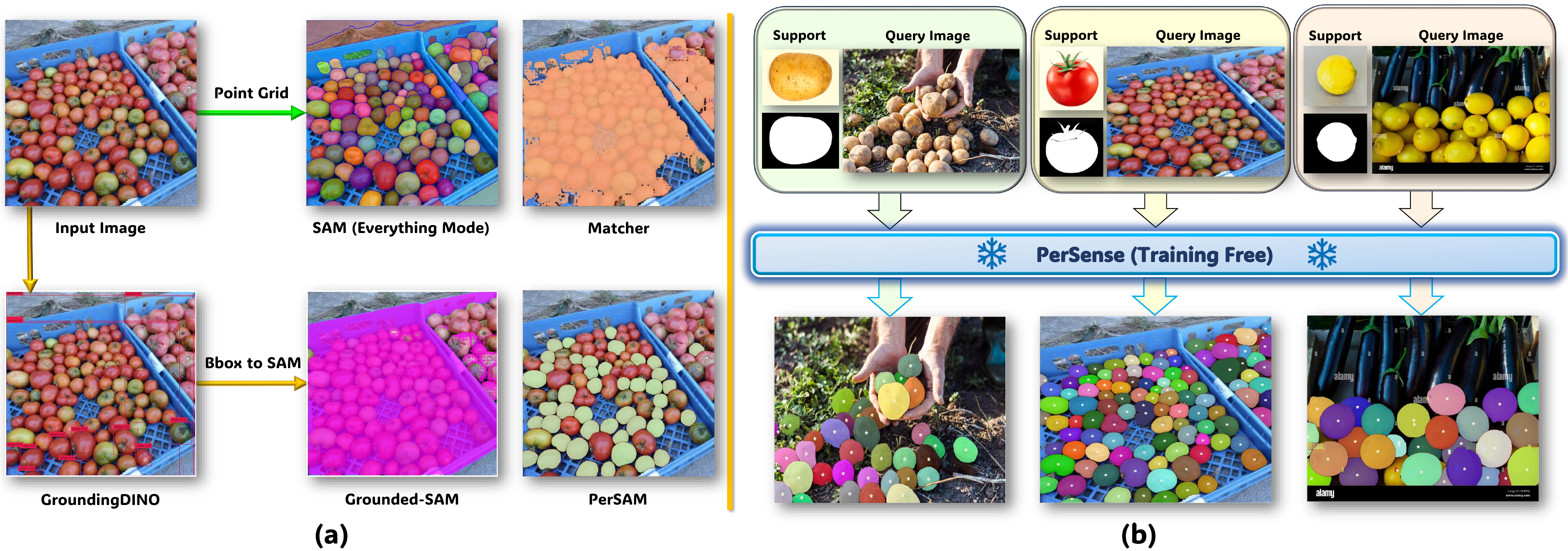}
    \caption{(a) Recent methods struggle in dense scenes, while SAM's “everything mode” segments all regions indiscriminately. (b) Introducing PerSense: a \textbf{training-free, model-agnostic one-shot framework} for personalized instance segmentation in dense images.}
    \label{fig:introfig}
\end{figure}

\vspace{-5pt}
\section{Introduction}
\label{sec:intro}
Imagine working in a food processing sector where the goal is to automate the quality control process for vegetables, such as potatoes, using vision sensors. The challenge is to segment all potato instances in densely packed environments, where variations in scale, occlusions, and background clutter add complexity to the task. We refer to this task as \textit{personalized instance segmentation in dense images}, building on the concept of personalized segmentation, first introduced in~\cite{zhang2023personalize}. The term \textit{personalized} refers to the segmentation of a specific visual category / concept within an image. Our task setting focuses on personalized instance segmentation, particularly in \textit{dense scenarios}.

A natural approach to address this problem is leveraging state-of-the-art (SOTA) foundation models. One key contribution is Segment Anything Model (SAM)~\cite{kirillov2023segment}, which introduces a prompt-driven segmentation framework. However, SAM lacks the ability to segment distinct visual concepts~\cite{zhang2023personalize}. Its "everything mode" uses a point grid to segment all objects, including both foreground and background (Fig.~\ref{fig:introfig}). Alternatively, users can provide manual prompts to isolate specific instances, making the process labor-intensive and impractical. One approach to automation is using box prompts from a pre-trained object detector to isolate the object of interest. Grounded-SAM~\cite{ren2024grounded} follows this strategy by forwarding bounding boxes from GroundingDINO~\cite{liu2023grounding} to SAM~\cite{kirillov2023segment} for segmentation. However, bounding boxes are limited by box shape, occlusions, and the orientation of objects~\cite{zand2021oriented}. A standard axis-aligned box for a particular object may include portions of adjacent instances. Additionally, when using non-max suppression (NMS), bounding box-based detections may group closely positioned instances of the same object together. Although techniques like bipartite matching introduced in DETR~\citep{carion2020end} can address the NMS issue, bounding box-based detections are still challenged due to variations in object scale, occlusions, and background clutter. These challenges become even more pronounced in dense scenes~\cite{wan2019adaptive}, making precise instance segmentation increasingly difficult (Fig.~\ref{fig:introfig}). Point-based prompting, mostly based on manual user input, is generally better than box-based prompting for tasks that require high accuracy, fine-grained control, and the ability to handle occlusions, clutter, and dense instances~\cite{maninis2018deep}. However, the automated generation of point prompts using low-shot data, for personalized segmentation in dense scenarios, has largely remained unexplored. Recent works, such as SegGPT~\cite{wang2023seggpt}, PerSAM~\cite{zhang2023personalize} and Matcher~\cite{liu2023matcher}, introduce frameworks for one-shot personalized segmentation. Despite their effectiveness in sparsely populated scenes with clearly delineated objects, these methods show limited performance in dense scenarios (Fig.~\ref{fig:introfig}).  

\begin{figure}[t]
    \centering
    \includegraphics[width=1\linewidth]{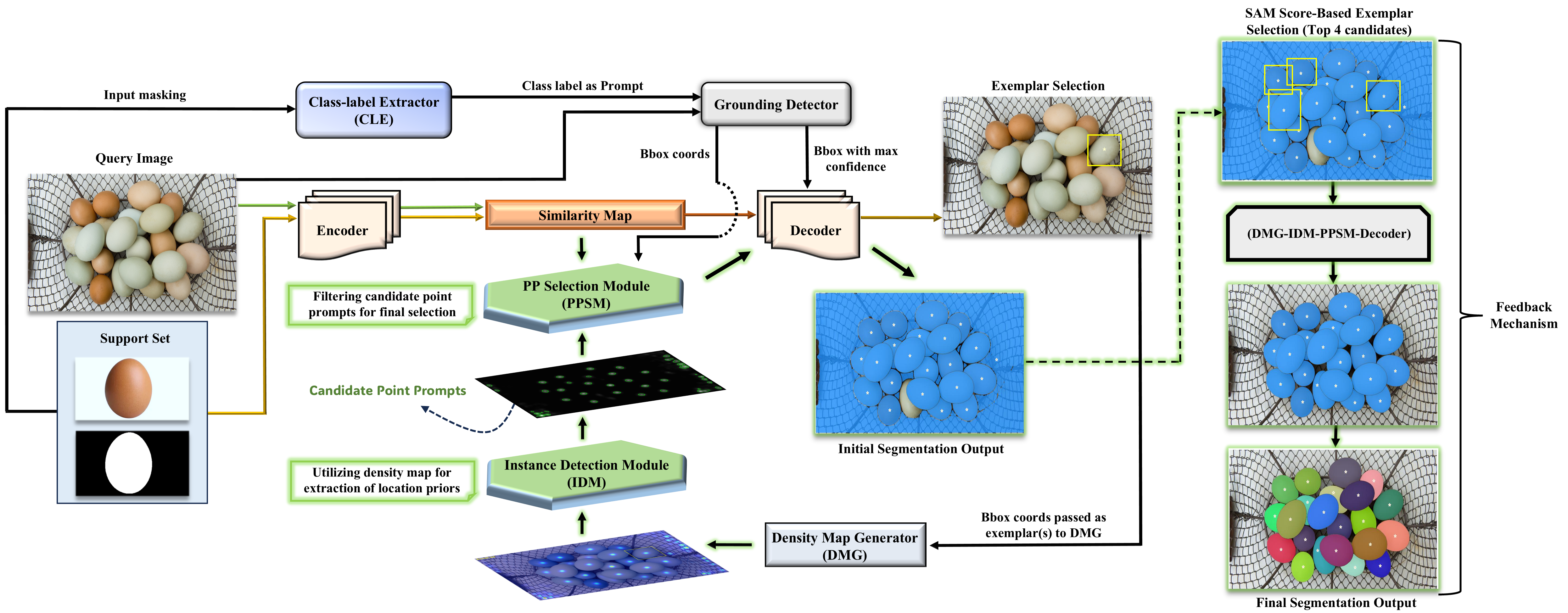}
    \caption{Overview: PerSense begins by extracting the object category using CLE and selecting exemplar based on similarity map and grounded detections. This exemplar is then forwarded to DMG for DM generation. IDM extracts candidate point prompts from DM, which are refined by PPSM to suppress false positives. The decoder then produces an initial segmentation mask, which is leveraged by feedback mechanism to select improved exemplars and regenerate an improved DM. This updated map enables IDM and PPSM to produce high-quality prompts, resulting in personalized instance segmentation in dense images.}
    \label{fig:mainfig}
\end{figure}

We approach this problem by exploring density estimation methods, which utilize density maps (DMs) to capture the spatial distribution of objects in dense scenes. While DMs effectively estimate global object counts, they struggle with precise instance-level localization~\cite{pelhan2024dave}. To this end, \textbf{\emph{we introduce PerSense}, an end-to-end, training-free and model-agnostic one-shot framework} (Fig.~\ref{fig:mainfig}), wherein we first develop a new baseline capable of autonomously generating instance-level candidate point prompts via a proposed Instance Detection Module (IDM), which exploits DMs for precise localization. We generate DMs using a density map generator (DMG) which highlights spatial distribution of object of interest based on input exemplars. To allow automatic selection of effective exemplars for DMG, we automate the process via a class-label extractor (CLE) and a grounding detector. Second, we design a Point Prompt Selection Module (PPSM) to mitigate false positives within the candidate point prompts using an adaptive threshold and box-gating mechanism. Both IDM and PPSM are plug-and-play components, seamlessly integrating into PerSense. Lastly, we introduce a robust feedback mechanism, which automatically  refines the initial exemplar selection by identifying multiple rich exemplars for DMG based on the initial segmentation output of PerSense. This ability to segment personalized concepts in dense scenarios is pivotal for industrial automation tasks such as quality control and cargo monitoring using vision-based sensors. Beyond industry, it holds promise for medical applications, particularly in cellular-level segmentation. Extensive experiments demonstrate that PerSense outperforms SOTA methods in both performance and efficiency for dense settings. Finally, to our knowledge, no existing benchmark specifically targets segmentation in dense images. While datasets like COCO~\cite{lin2014microsoft}, LVIS~\cite{gupta2019lvis}, and FSS-1000~\cite{li2020fss} include multi-instance images, they lack dense scenarios due to limited object counts. For instance, LVIS averages only 3.3 instances per category. To bridge this gap, \textit{we introduce PerSense-D, a one-shot segmentation benchmark for dense images. It comprises 717 images across 28 object categories, with an average of 53 instances per image, along with dot annotations and ground truth masks}. Featuring heavy occlusion and background clutter, PerSense-D offers both class-wise and density-based categorization for fine-grained evaluation in real-world dense settings.

% Finally, to our knowledge, no benchmark specifically targets segmentation in dense images. While mainstream datasets like COCO\cite{lin2014microsoft}, LVIS\cite{gupta2019lvis}, and FSS-1000~\cite{li2020fss} contain images with multiple instances of the same object, they do not represent dense scenarios due to their limited object count. For example, images in the LVIS dataset contain an average of 11.2 instances across 3.4 object categories (3.3 instances per category), which is insufficient to represent dense scenarios. To bridge this gap, we introduce PerSense-D, a personalized one-shot evaluation benchmark designed exclusively for segmentation in dense images.  PerSense-D consists of 717 images across 28 diverse object categories, with an  \textit{average of 53 object instances per image}. It includes dot annotations and ground truth masks, featuring images with significant occlusion and background clutter, making it a unique and challenging benchmark for advancing algorithmic development and practical tools in dense environments. Beyond class-wise division, it also provides density-based categorization for fine-grained evaluation. 
\vspace{-3pt}
\section{Related Work}
\label{sec:relatedwork}

\noindent\textbf{One-shot Personalized Segmentation:}  
As discussed in Sec.~\ref{sec:intro}, SAM~\cite{kirillov2023segment} lacks semantic awareness, limiting its ability to segment personalized concepts. PerSAM~\cite{zhang2023personalize} addresses this by introducing a training-free one-shot framework using iterative masking. However, in dense scenes, its performance degrades due to (1) high computational cost from iteration scaling, (2) declining confidence map quality with more masked objects, and (3) premature termination from fixed confidence thresholds. In contrast, PerSense uses DMs to generate instance-level point prompts in a single pass, improving efficiency and accuracy in dense settings. SLiMe~\cite{khani2023slime} enables segmentation based on granularity in the support set, but struggles with small objects due to limited attention resolution in Stable Diffusion~\cite{rombach2022high}. Painter~\cite{wang2023images} employs masked image modeling with in-context prompting, but focuses on coarse global features, limiting fine-grained segmentation. SegGPT~\cite{wang2023seggpt} improves generalization via random coloring scheme but oversimplifies dense regions under one-shot settings, making object separation difficult. Matcher~\cite{liu2023matcher} combines feature extraction with bidirectional matching, but its reliance on clustering (for instance sampling) and box prompt limits instance-level performance in dense scenes. PerSense avoids clustering and sampling altogether by using DMG to generate personalized DMs and directly produce precise point prompts, making it more effective in complex dense environments.

\noindent\textbf{Interactive segmentation:} Recent methods like InterFormer~\cite{Huang_2023_ICCV}, MIS~\cite{Li_2023_ICCV}, SEEM~\cite{zou2024segment}, and Semantic-SAM~\cite{li2024segment} support interactive segmentation with semantic awareness or multi-granularity refinement, but all rely on manual prompts, limiting scalability. Moreover, they do not explicitly generalize from a one-shot reference to all instances of the same category within an image. In contrast, PerSense automates instance-specific point prompt generation from a single reference, enabling personalized segmentation without manual intervention.

% \noindent\textbf{Interactive segmentation:} 
% Recently, interactive segmentation has gained attention, with models like InterFormer~\cite{Huang_2023_ICCV}, MIS~\cite{Li_2023_ICCV}, and SEEM~\cite{zou2024segment} offering user-friendly interfaces but relying on manual input from the user, which limits scalability. More recently,  Semantic-SAM~\cite{li2024segment} improves upon vanilla SAM by incorporating semantic-awareness and multi-granularity segmentation, however it still requires manual prompts and does not explicitly generalize from one-shot reference to all instances of the same category within the image. In contrast, PerSense automates instance-specific point prompt generation from one-shot data, enabling personalized segmentation without manual intervention.

\section{Methodology}
\label{sec:method}
% Fig.~\ref{fig:mainfig} illustrates PerSense overview. See supplementary S1 and S2 for pseudo codes and component-wise analysis.

\begin{figure}[t]
    \centering
    \includegraphics[width=0.8\linewidth]{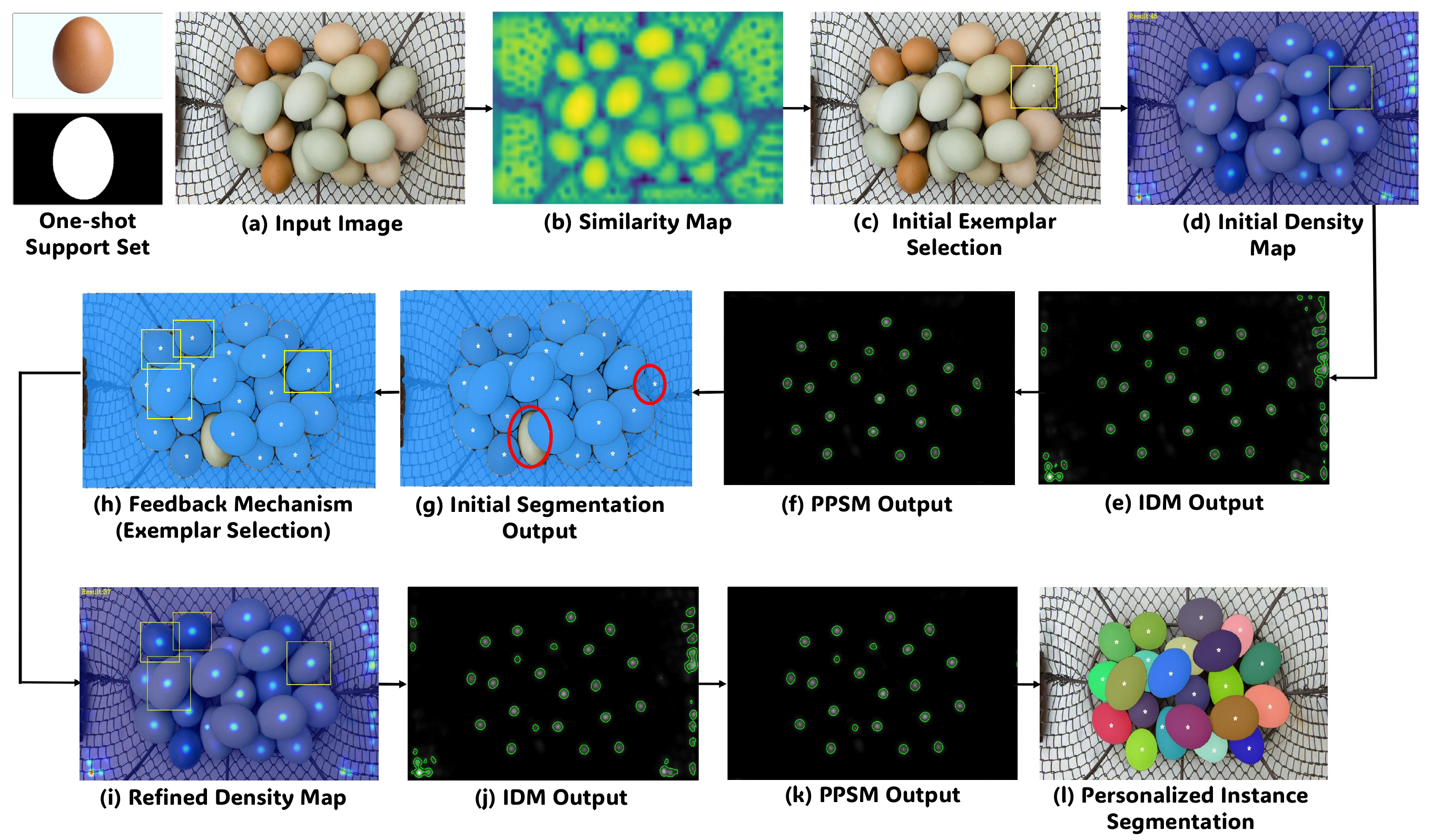}
    \caption{Step-by-step workflow of PerSense components. From an input image (a), a similarity map (b) is generated using the support set. Exemplar selection (c), guided by similarity scores and grounded detections, produces an initial density map (DM) (d) via DMG.  The IDM processes this DM to generate candidate point prompts (e), which are refined by PPSM (f) to filter false positives. The decoder yields an initial segmentation based on refined prompts (g); however, a few false positive prompts still remain alongside false negatives (red circle). The feedback mechanism (h) leverages initial segmentation to improve exemplar selection based on SAM scores, generating an improved DM (i). IDM and PPSM extracts refined prompts (j, k) from improved DM leading to the final segmentation output (l).}
    \label{fig:stepwise}
\end{figure}

We introduce PerSense, a training-free and model-agnostic one-shot framework for personalized instance segmentation in dense images (Fig.~\ref{fig:mainfig}). In the following sections, we detail the core components of our PerSense framework. The step-by-step working of PerSense components is depicted in Fig.~\ref{fig:stepwise}. See Appendix (\ref{algorithm}) for pseudo-codes.

\noindent\textbf{Class-label Extraction and Exemplar Selection:} PerSense operates as a one-shot training-free framework (Fig.~\ref{fig:mainfig}), leveraging a support set to guide personalized segmentation in a query image based on semantic similarity with the support object. First, input masking is applied to the support image using a coarse support mask to isolate the object of interest. The masked image is then processed by the CLE with the prompt, "\textit{Name the object in the image?}". The CLE generates a description, from which the noun is extracted as the object category. This category serves as a prompt for the grounding detector, enabling personalized object detection in the query image. To refine the prompt, the term "all" is prefixed to the class label. Next, we compute the cosine similarity score \( S_{score} \) between query \( Q \) and support \( S_{supp} \) features, extracted by the encoder, and locate the pixel-precise point \( P_{max} \) with the highest similarity score within the most confident bounding box \( B_{max} \):
\vspace{-2pt}
\begin{equation}
S_{score}(Q, S_{supp}) = \text{cos\_sim}(f(Q), f(S_{supp})), \quad P_{max} = \arg\max_{P \in B_{max}} S_{score}(P, S_{supp}),
\end{equation}where \( f(\cdot) \) represents the encoder and \( P \) denotes candidate points within the bounding box \( B_{max} \). The identified point serves as the positive location prior, which is subsequently fed to the decoder for segmentation. Additionally, we extract the bounding box surrounding the segmentation mask of the object. This bounding box is then forwarded as an exemplar to the DMG for generation of the DM.

\begin{figure}[!t]
    \centering
    \includegraphics[width=0.9\linewidth]{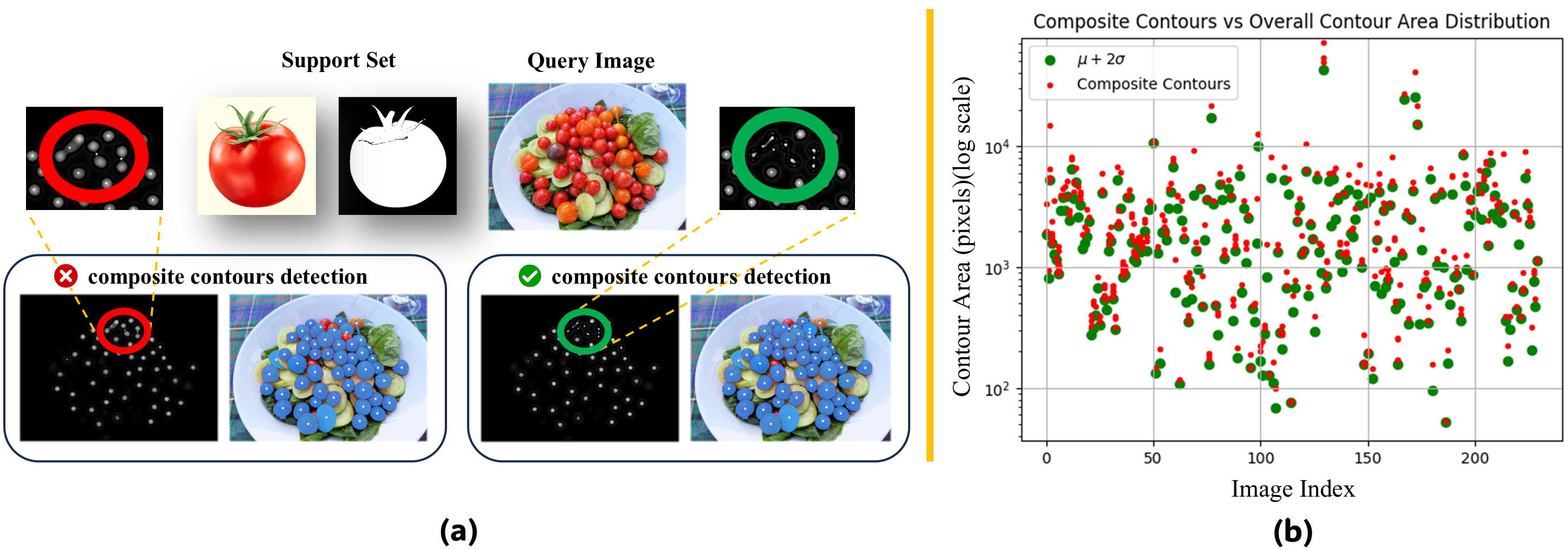}
    \caption{(a) Without identifying composite contours (red circle). With identification of composite contours (green circle) [best viewed in zoom]. (b) Contour area distribution across 250 dense images, highlighting composite contours as outliers beyond \( \mu + 2\sigma \).}
    \label{fig:comp_contour} \vspace{-3pt}
\end{figure}

\vspace{1pt}
\noindent\textbf{Instance Detection Module (IDM):} The IDM begins by converting the DM from the DMG into a grayscale image \( I_{gray} \). A binary image \( I_{binary} \) is created from \( I_{gray} \) using a pixel-level threshold \( T \) (\( T \in [0, 255] \)), followed by a morphological erosion operation using a \( 3 \times 3 \) kernel \( K \):
\begin{equation}
I_{binary}(x, y) = 
\begin{cases} 
1 & \text{if } I_{gray}(x, y) \geq T \\
0 & \text{if } I_{gray}(x, y) < T 
\end{cases},
\quad
I_{eroded}(x, y) = \min_{(i,j) \in K} I_{binary}(x+i, y+j),
\end{equation}
where \( I_{eroded} \) is the eroded image, and \( (i,j) \) iterates over the kernel \( K \) to refine the boundaries and eliminate noise from the binary image. A small kernel is deliberately used to avoid damaging the original densities of true positives. Next, contours are extracted from \( I_{eroded} \), and their areas \( A_\text{ctr} \sim \mathcal{N}(\mu, \sigma^2) \) are modeled as a Gaussian distribution, where \( \mu \) represents the mean contour area corresponding to the typical object size, and \( \sigma \) denotes the standard deviation, capturing variations in contour areas due to differences in object sizes among instances. Composite contours, which encapsulate multiple objects within a single contour, are identified using a threshold \( T_\text{comp} \), defined as \( \mu + 2\sigma \) based on the contour size distribution (Fig.~\ref{fig:comp_contour}). These regions, though rare, are detected as outliers exceeding \( T_\text{comp} \).  The mean, standard deviation, and probability of a contour being composite are computed as:
\begin{equation}
\mu = \frac{1}{N} \sum_{i=1}^N A_i, \quad \sigma = \sqrt{\frac{1}{N} \sum_{i=1}^N (A_i - \mu)^2}, \quad P(A_\text{ctr} > T_\text{comp}) = 1 - \Phi\left(\frac{T_\text{comp} - \mu}{\sigma}\right)
\end{equation}
where \( N \) is the number of detected contours, and \( \Phi \) is the cumulative distribution function (CDF) of the standard normal distribution. For each composite contour, a distance transform is applied to reveal internal sub-regions representing individual object instances, defined as \( D_\text{transform}(x, y) = \min_{(i, j) \in B} \| (x, y) - (i, j) \| \), where \( B \) represents contour boundary pixels and \( (x, y) \) are the coordinates of each pixel within the region of interest. A binary threshold applied to \( D_\text{transform} \) segments sub-regions within each composite contour, enabling separate identification of overlapping objects in dense scenarios. For each detected contour (including both parent and child), the centroid is computed using spatial moments to account for the irregular and non-uniform shapes of the contours.
\begin{equation}
cX = \frac{M_{10}}{M_{00} + \epsilon}, \quad cY = \frac{M_{01}}{M_{00} + \epsilon}, \quad (M_{00}, M_{10}, M_{01}) = \sum_x \sum_y I(x, y) \cdot (1, x, y)
\end{equation}
where \( M_{pq} \) are the spatial moments, \( \epsilon \) is a small constant to prevent division by zero, and \( I(x, y) \) is the pixel intensity at position \( (x, y) \). These centroids serve as candidate point prompts, accurately marking the locations of individual object instances in dense scenarios. The candidate point prompts are subsequently forwarded to PPSM for final selection.

 % As \( C \) increases, \( T_\text{adapt} \) decreases, resulting in a more inclusive threshold that admits additional candidate points under high-density conditions.

% \subsection{Point Prompt Selection Module (PPSM)}
% \label{subsection:PPSM}
\vspace{1pt}
\noindent\textbf{Point Prompt Selection Module (PPSM):} PPSM is a key component in PerSense, filtering candidate point prompts before passing to the decoder. Each candidate point from IDM is evaluated based on its query-support similarity score, using an adaptive threshold that adjusts dynamically to object density. We statistically model the adaptive threshold in PPSM, scaling it based on object count. Assuming similarity scores \( S(x, y) \) follow a Gaussian distribution with mean \( \mu \) and variance \( \sigma^2 \), we define the adaptive threshold \( T_\text{adapt} \) as:
\vspace{-3pt}
\begin{equation}
T_\text{adapt} = \frac{S_{\text{max}}}{C / k}, \quad \text{for } C > 1
\end{equation}
\vspace{-1pt}where \( S_\text{max} \) is the maximum similarity score, representing the most aligned point with the target feature, and \( C \) and \( k \) denote the object count and a normalization constant, respectively. When \( C = 1 \), the point associated with \( S_\text{max} \) is selected as the prompt. We choose \( k = \sqrt{2} \), based on empirical results presented in Sec.~\ref{sec:experiments}. The probability \( P \) of a randomly selected point exceeding this adaptive threshold is given by:
\vspace{-2pt}
\begin{equation}
P(S \geq T_\text{adapt}) = 1 - \Phi\left(\frac{T_\text{adapt} - \mu}{\sigma}\right) = 1 - \Phi\left(\frac{\frac{S_{\text{max}}}{C / k} - \mu}{\sigma}\right)
\end{equation}
\vspace{-1pt}where \( \Phi \) is the CDF of standard normal distribution. As \( T_\text{adapt} \) becomes more permissive with increasing \( C \), the probability of candidate points surpassing the threshold also rises, ensuring greater flexibility in dense scenarios. This density-aware adjustment mitigates the risk of inadvertently excluding true positives, a limitation commonly observed with fixed thresholds tuned primarily to suppress false positives. Such adaptability is crucial, as similarity scores can vary significantly due to minor intra-class differences. In highly dense images (\( C > 50 \)), the score distribution widens due to increased intra-class variability, making dynamic thresholding essential for robust and reliable prompt selection.  To enhance spatial precision, we also introduce a box gating mechanism: a point is retained only if (i) its similarity score exceeds \( T_\text{adapt} \), and (ii) it falls within at least one of the boxes predicted by the grounding detector. This ensures selected prompts are both semantically aligned and spatially grounded.

\vspace{1pt}
\noindent\textbf{Feedback Mechanism:}  
PerSense introduces a feedback mechanism to improve exemplar selection for DMG by leveraging the initial segmentation mask (\( M_{seg} \)) and the corresponding SAM-generated mask scores (\( S_{mask} \)). The top \( m \) candidates are selected as exemplars using \( C_{Top} = \text{Top}_m(M_{seg}, S_{mask}, k) \), where \( C_{Top} \) denotes the top \( m \) masks ranked by SAM score. We set \( m = 4 \) (see Sec.~\ref{sec:experiments}). These candidates are then forwarded to DMG in a feedback manner, enhancing the quality of the DM and, consequently, the segmentation performance.

% A quantitative analysis of this mechanism is presented in Sec.~\ref{sec:experiments}.

% \vspace{3pt}
% \noindent\textbf{Feedback Mechanism:} PerSense proposes a feedback mechanism to enhance the exemplar selection process for the DMG by leveraging the initial segmentation mask (\( M_{seg} \)) from the decoder and the corresponding SAM-generated mask scores (\( S_{mask} \)).
% \vspace{-3pt}
% \begin{equation}
% C_{Top} =\text{Top}_k(M_{seg}, S_{mask}, k),
% \end{equation}
% where \( C_{Top} \) represents the set of the top \( k \) candidates, selected based on their mask scores. In our case  \( k = 4 \)  (see Sec.~\ref{subsection:ablation_study}). These selected candidates are then forwarded as exemplars to DMG in a feedback manner. This leads to improved accuracy of the DM leading to enhanced segmentation performance. The quantitative analysis of this aspect is further discussed in Sec.~\ref{sec:experiments}, which explicitly highlights the value added by the proposed feedback mechanism.
\begin{figure}[t]
    \centering
    \includegraphics[width=1\linewidth]{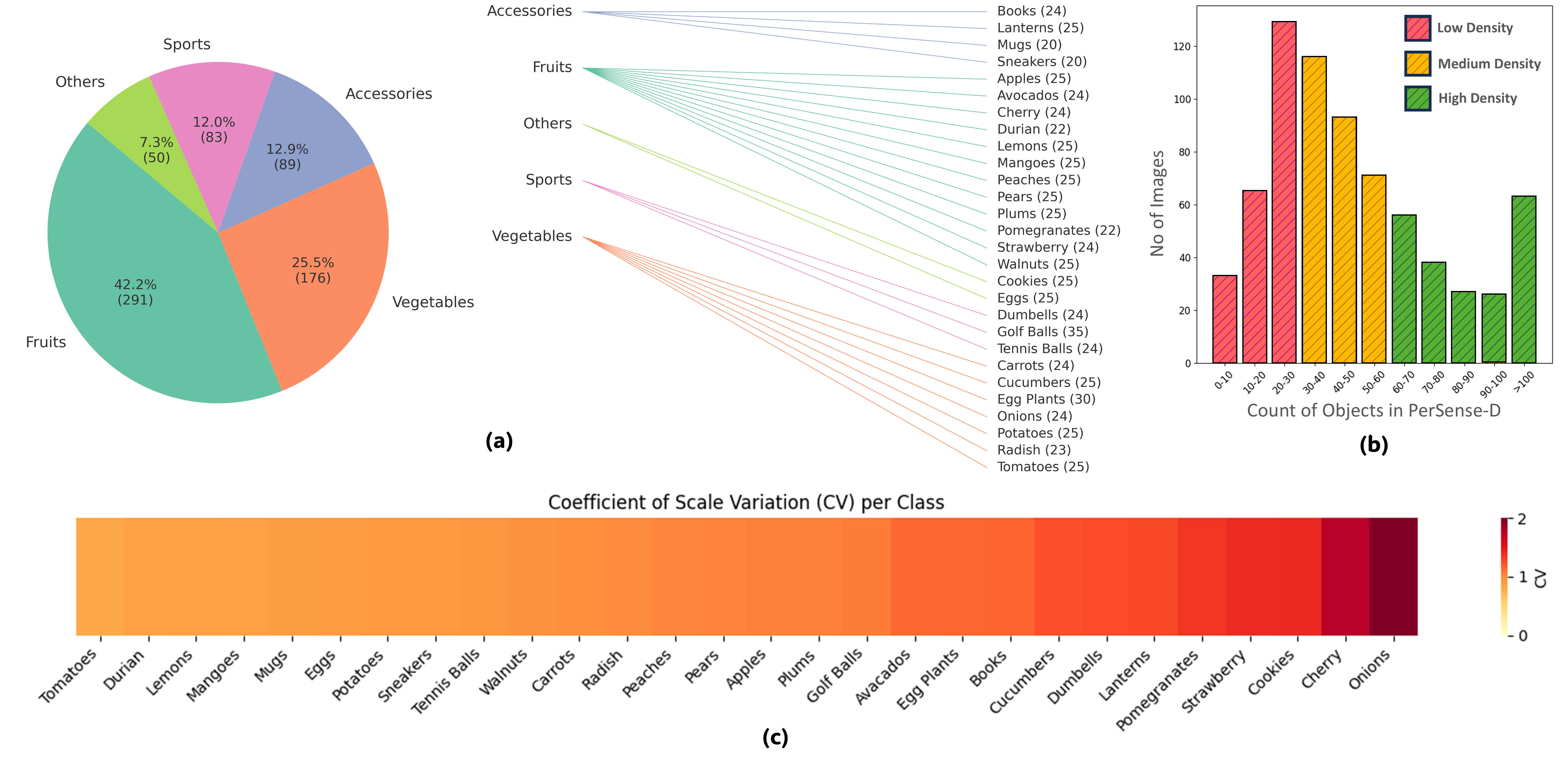}
    \caption{(a) Object categories in PerSense-D. (b) Image distribution across object count bins grouped by density levels. (c) Coefficient of Variation (CV) in object scale across classes, defined as CV = $\sigma / \mu$. Higher values indicate greater intra-class scale variability.}
    \label{fig:dtasetstats}
    \vspace{-4pt}
\end{figure}

\vspace{-4pt}
\section{New Evaluation Benchmark (PerSense-D)}
\label{sec:benchmark}

% PerSense leverages DMs from DMG for point prompt extraction via IDM and PPSM, specifically targeting dense images. 
% \vspace{1pt}
% \noindent\textbf{New Evaluation Benchmark (PerSense-D):} 
Existing segmentation datasets like COCO~\cite{lin2014microsoft}, LVIS~\cite{gupta2019lvis}, and FSS-1000~\cite{li2020fss} include multi-instance images but rarely represent dense scenes due to limited object counts. For example, LVIS averages only 3.3 instances per category. To bridge this gap, we introduce PerSense-D, \textit{a dense segmentation benchmark of 717 images spanning across 28 object categories}   (Fig.~\ref{fig:dtasetstats}). On average, each image contains 53 object instances, with a total of 36,837 annotated objects across the entire dataset. The number of instances per image ranges from 7 to 573, and the average image resolution is 839 \(\times\) 967 pixels. In addition to dense layouts, PerSense-D exhibits high intra-class object scale diversity, as reflected by the coefficient of variation (CV) in object scale across categories (Fig.~\ref{fig:dtasetstats}). This enhances its value as a challenging testbed for advancing segmentation methods in real-world dense scenarios.

\noindent\textbf{\textit{Image Collection and Retrieval:}} To collect the dense set, we searched for object categories across Google, Bing, and Yahoo, using prefixed keywords like "\textit{multiple}", "\textit{lots of}", and "\textit{many}" to encourage dense results. For each of the 28 categories, we retrieved the top 100 images, yielding 2800 candidates for further filtering.

% \noindent\textbf{Image Collection and Retrieval:}
% Out of 717 images, we have 689 dense query images and 28 support images. To acquire the set of 689 dense images, we initiated the process with a collection of candidate images obtained through keyword searches. To mitigate bias, we retrieved the candidate images by querying object keywords across three distinct Internet search engines: Google, Bing, and Yahoo. To diversify the search query keywords, we prefixed adjectives such as  '\textit{multiple}', '\textit{lots of}', and '\textit{many}' before the category names. In every search, we collected the first 100 images. With 28 categories, we gathered a total of 2800 images, which were subsequently filtered in the next step.

%Object categories in PerSense-D are selected as a subset of Objects365 dataset~\cite{shao2019objects365}. This was done to ensure fair comparison with Grounded-SAM~\cite{ren2024grounded}, as its component model GroundingDINO~\cite{liu2023grounding} is pre-trained on Objects365 dataset.

\noindent\textbf{\textit{Manual Inspection and Filtering:}}  
Candidate images were manually filtered based on the following criteria: (1) Adequate quality for clear object distinction, (2) at least 7 object instances per image, following dense counting datasets~\cite{ranjan2021learning}, and (3) presence of dense, cluttered scenes with occlusions. This resulted in 689 images from an initial pool of 2800.

% \noindent\textbf{Manual Inspection and Filtering:}
% The candidate images were manually inspected following a three-point criterion. (1) The image quality and resolution should be sufficiently high to enable easy differentiation between objects. (2) Following the criterion in dense counting datasets such as  FSC-147~\cite{ranjan2021learning}, we set the minimum object count to 7 per image for our PerSense-D benchmark. (3) The image shall contain a challenging dense environment with sufficient occlusions among object instances along with background clutter. Based on this criterion, we filtered 689 out of 2800 images.

\noindent\textbf{\textit{Semi-automatic Image Annotation Pipeline:}}  
We crowdsourced annotations using a semi-automatic pipeline. Following a model-in-the-loop approach~\cite{kirillov2023segment}, PerSense generated initial masks, which annotators refined using OpenCV and Photoshop tools for pixel-accurate corrections. Given the high instance count,  manual refinement averaged 15 minutes per image. We also provide dot annotations at object centers. See Appendix (\ref{annotated_images}) for examples.

\noindent\textbf{\textit{Evaluation Protocol:}} To ensure fairness under one-shot setting, we provide a set of 28 support images (labeled “00”), each containing a single object instance per category. This eliminates randomness in support selection and standardizes evaluation across methods. PerSense-D supports both class-wise and density-based evaluations for granular performance analysis. For the latter, images are categorized into three density levels based on object count: \textit{Low} (\(C_I \leq 30\), 224 images), \textit{Medium} (\(30 < C_I \leq 60\), 266 images), and \textit{High} (\(C_I > 60\), 199 images). Fig.~\ref{fig:dtasetstats} shows image distribution by object count range and density category.

\vspace{-9pt}

\section{Experiments and Ablations}
\label{sec:experiments}

% \subsection{Implementation Details}
% \label{subsection:implementation_details}

\noindent\textbf{Implementation Details:} We use VIP-LLaVA~\citep{cai2024vipllava} as the CLE, built on CLIP-336px~\citep{radford2021learning} and Vicuna v1.5~\citep{chiang2023vicuna}. For grounding, we adopt GroundingDINO~\citep{liu2023grounding} with a detection threshold of 15\%. To demonstrate model-agnostic capability of PerSense, we use DSALVANet~\citep{he2024few} and CounTR~\citep{liu2022countr}, both pretrained on FSC-147~\citep{ranjan2021learning}, as DMG1 and DMG2, respectively. For fair comparison, we use vanilla SAM~\citep{kirillov2023segment} encoder and decoder. PerSense is evaluated using the standard mIoU metric on PerSense-D, COCO~\cite{lin2014microsoft}, and LVIS~\cite{gupta2019lvis} datasets. \textbf{\textit{No training is involved in any of our experiments.}}

\begin{figure}[t]
    \centering
    \includegraphics[width=1\linewidth]{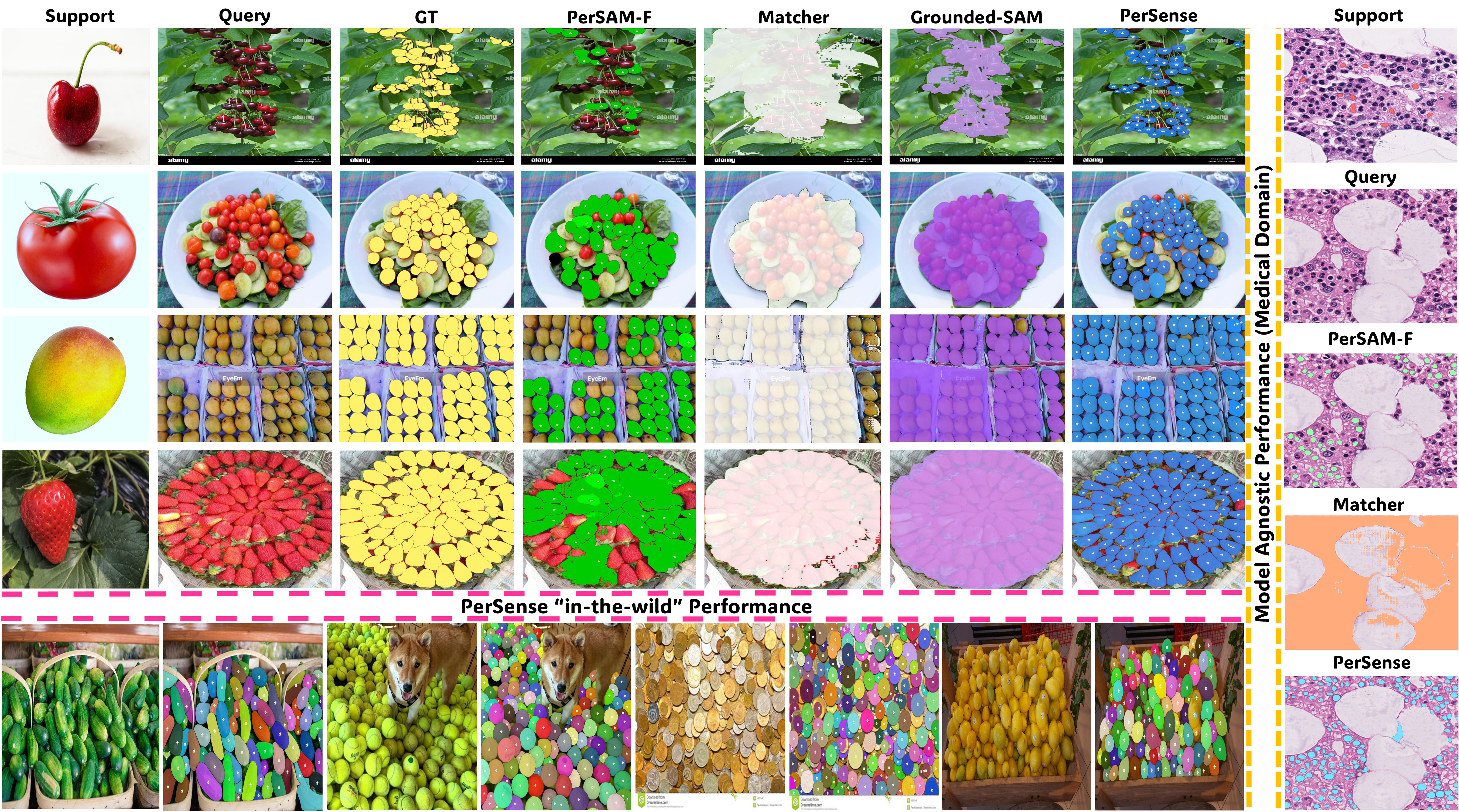}
    \caption{Qualitative comparison of PerSense with SOTA on PerSense-D. Last row and last column illustrate \textit{in-the-wild} and \textit{model-agnostic} performance, respectively.}
    \label{fig:qualitative_results}
    \vspace{-8pt}
\end{figure}

\noindent\textbf{Performance on Dense Benchmark:}  
Table~\ref{comparisonwithpersensed} shows that PerSense achieves an overall class-wise mIoU of 71.61\% on PerSense-D, outperforming one-shot segmentation methods including PerSAM~\cite{zhang2023personalize} (+47.16\%), PerSAM-F (+42.27\%), SegGPT~\cite{wang2023seggpt} (+16.11\%), Matcher~\cite{liu2023matcher} (+8.83\%), and Grounded-SAM (+5.69\%). We also compare PerSense with dense object counting methods: it surpasses the training-free TFOC~\cite{shi2024training} by +8.98\%, and outperforms PseCo~\cite{huang2024point}, GeCo~\cite{pelhannovel}, and C3Det~\cite{lee2022interactive} by +9.78\%, +5.66\%, and +23.01\%, respectively. Fig.~\ref{fig:qualitative_results} showcases qualitative results on PerSense-D along with in-the-wild examples. A Class-wise comparison is presented in Appendix (\ref{classmiou}). For a fine-grained analysis, we also evaluate PerSense across low, medium, and high-density categories, where it consistently outperforms SOTA (Table~\ref{comparisonwithpersensed}). Notably, performance improves monotonically from low to high density, reflecting PerSense's strength in dense scenes.

\begin{table*}[t]
\centering
\setlength{\tabcolsep}{2pt}
\fontsize{6.2pt}{6.2pt}\selectfont

\begin{minipage}[t]{0.475\textwidth}
\centering
\begin{tabular}{llc|c|c@{\hskip 1pt}c} 
\toprule
\multirow{3}{*}{\textbf{Method}} & \multirow{3}{*}{\textbf{Venue}} 
& \multicolumn{3}{c}{\textbf{\shortstack{Density-based\phantom{\textsuperscript{i}}}}} 
& \multicolumn{1}{c}{\textbf{\shortstack{Class-wise\phantom{\textsuperscript{i}}}}} \\ 
\cmidrule(lr){3-5} \cmidrule(lr){6-6}
& & \textbf{Low} & \textbf{Med} & \textbf{High} & \textbf{\shortstack{Overall}} \\
\midrule
\multicolumn{6}{l}{\underline{\textit{End-to-end trained / fine-tuned}}} \vspace{5pt} \\
C3Det~\citep{lee2022interactive} & CVPR 2022 & 52.70 & 46.64 & 39.11 & 48.60 \\
SegGPT~\citep{wang2023seggpt} & ICCV 2023 & 59.81 & 53.34 & 52.05 & 55.50 \\
PerSAM-F$^\dagger$~\citep{zhang2023personalize} & ICLR 2024 & 38.18 & 34.84 & 26.73 & 29.30 \\
PseCo$^\dagger$~\citep{huang2024point} & CVPR 2024 & 53.99 & 65.23 & 68.55 & 61.83 \\
GeCo$^\dagger$~\citep{pelhannovel} & NIPS 2024 & 63.92 & 63.40 & 74.49 & 65.95 \\
\midrule
\multicolumn{6}{l}{\underline{\textit{Training-free ( $^\dagger$ indicates methods using SAM)}}} \vspace{3pt}\\
PerSAM$^\dagger$~\citep{zhang2023personalize} & ICLR 2024 & 32.27 & 28.75 & 20.25 & 24.45 \\
TFOC$^\dagger$~\citep{shi2024training} & WACV 2024 & \underline{62.78} & 65.38 & 65.69 & 62.63 \\
Matcher$^\dagger$~\citep{liu2023matcher} & ICLR 2024 & 58.62 & 58.30 & 68.00 & 62.80 \\
GroundedSAM$^\dagger$~\citep{ren2024grounded} & arXiv 2024 & 58.36 & 66.24 & 64.97 & 65.92 \\
\textbf{PerSense}$^\dagger$ (DMG1) & (ours) & \textbf{66.36} & \underline{67.27} & \underline{74.78} & \underline{70.96} \\
\textbf{PerSense}$^\dagger$ (DMG2) & (ours) & 59.84 & \textbf{73.51} & \textbf{77.57} & \textbf{71.61} \\
\bottomrule
\end{tabular}
\vspace{1pt}
\caption{Comparison of PerSense with SOTA on PerSense-D. Density-based and class-wise mIoU are reported.}
\label{comparisonwithpersensed}
\vspace{-2pt}
\end{minipage}
\hfill
\begin{minipage}[t]{0.475\textwidth}
\centering
\begin{tabular}{llc|c|c|c|c@{\hskip 1pt}c} 
\toprule
\multirow{3}{*}{\textbf{Method}} & \multirow{3}{*}{\textbf{Venue}} & \multicolumn{5}{c}{\textbf{COCO-20\textsuperscript{i}}} & \textbf{LVIS-92\textsuperscript{i}} \\ 
 \cmidrule(lr){8-8}
& & \textbf{F0} & \textbf{F1} & \textbf{F2} & \textbf{F3} & \textbf{\shortstack{Mean}} & \textbf{\shortstack{Mean}} \\ 
\midrule
\multicolumn{8}{l}{\underline{\textit{In-domain training}}} \vspace{1pt} \\
HSNet~\citep{min2021hypercorrelation} & CVPR 21 & 37.2 & 44.1 & 42.4 & 41.3 & 41.2 & 17.4 \\
VAT~\citep{hong2022cost} & ECCV 22 & 39.0 & 43.8 & 42.6 & 39.7 & 41.3 & 18.5 \\
FPTrans~\citep{zhang2022feature} & NIPS 22 & 44.4 & 48.9 & 50.6 & 44.0 & 47.0 & - \\
MIANet~\citep{yang2023mianet} & CVPR 23 & 42.4 & 52.9 & 47.7 & 47.4 & 47.6 & - \\
LLaFS~\citep{zhu2024llafs} & CVPR 24 & 47.5 & 58.8 & 56.2 & 53.0 & 53.9 & - \\
\midrule
\multicolumn{8}{l}{\underline{\textit{COCO as training data}}} \vspace{1pt}\\
Painter~\citep{wang2023images} & CVPR 23 & 31.2 & 35.3 & 33.5 & 32.4 & 33.1 & 10.5 \\
SegGPT~\citep{wang2023seggpt} & ICCV 23 & 56.3 & 57.4 & 58.9 & 51.7 & 56.1 & 18.6 \\
\midrule
\multicolumn{8}{l}{\underline{\textit{Training-free (excluding PerSAM-F~\citep{zhang2023personalize}) }}} \vspace{1pt}\\
PerSAM$^\dagger$~\citep{zhang2023personalize} &ICLR 24  & 23.1 & 23.6 & 22.0 & 23.4 & 23.0 & 11.5 \\
PerSAM-F$^\dagger$~\citep{zhang2023personalize} & ICLR 24 & 22.3 & 24.0 & 23.4 & 24.1 & 23.5 & 12.3 \\
Matcher$^\dagger$~\citep{liu2023matcher} &ICLR 24  & \textbf{52.7} & \textbf{53.5} & \textbf{52.6} & \textbf{52.1} & \textbf{52.7} & \textbf{33.0} \\
\textbf{PerSense}$^\dagger$ & (ours) &  \underline{47.8} &  \underline{49.3} &  \underline{48.9} &  \underline{50.1} &  \underline{49.0} &  \underline{25.7} \\ 
\bottomrule
\end{tabular}
\vspace{1pt}
\caption{Comparison of PerSense with SOTA approaches on COCO-20\textsuperscript{i} and LVIS-92\textsuperscript{i} in terms of mIoU.}
\label{Comparisonwithcoco}
\vspace{-2pt}
\end{minipage}
% \caption{Side-by-side comparison of PerSense with SOTA segmentation methods on PerSense-D (left) and other benchmarks (right).}
% \label{tab:persense_sidebyside}
\end{table*}

\begin{table*}[t] 
  \centering
  \fontsize{6.5pt}{6.5pt}\selectfont

  %---------------------- First Row: 3 Tables ----------------------
 \begin{minipage}[t]{0.32\textwidth}
  \centering
  \setlength{\tabcolsep}{2pt}
  \begin{tabular}{c c c}
    \toprule
    \multirow{3}{*}{\textbf{Method}} & \textbf{Memory} & \textbf{Avg Inf} \\
     & \multirow{2}{*}{\textbf{(MB)}} & \textbf{Time (sec)}\\[-1pt]
     & & {\scriptsize(\( C \) is count)} \\
    \midrule
    Grounded-SAM~\citep{ren2024grounded} & 2943 & 1.8 \\ 
    PerSAM~\citep{zhang2023personalize} & 2950 & \(( C \times 1.02)\) \\
    Matcher~\citep{liu2023matcher} & 3209 & 10.2 \\ 
    PerSense (ours) & 2988 & 2.7 \\ 
    \bottomrule
  \end{tabular}
  \vspace{2pt}
  \label{runningefficiency}
\end{minipage}%
  \hfill
  \begin{minipage}[t]{0.36\textwidth} 
    \centering
    \setlength{\tabcolsep}{2pt}
    \begin{tabular}{lccc}
      \toprule
       \multirow{2}{*}{\textbf{Modules}} & \textbf{Baseline} & \textbf{+ PPSM} & \textbf{PerSense} \\
       & \textbf{(mIoU)} & \textbf{(mIoU)} & \textbf{(mIoU)} \\
      \midrule
      IDM & yes & yes & yes \\
      PPSM & no & yes & yes \\ 
      Feedback & no & no & yes \\
      \midrule
      PerSense-D \textcolor{blue}{(Gain)} & 65.58 \textcolor{blue}{(-)} & 66.95 \textcolor{blue}{(+1.37)} & 70.96 \textcolor{blue}{(+4.01)} \\
      COCO \textcolor{blue}{(Gain)} & 46.33 \textcolor{blue}{(-)} & 48.81 \textcolor{blue}{(+2.48)} & 49.00 \textcolor{blue}{(+0.19)} \\
      \bottomrule
    \end{tabular}
    \vspace{2pt}
    \label{tab:component-wise}
  \end{minipage}%
  \hfill
  \begin{minipage}[t]{0.25\textwidth} 
    \centering
    \setlength{\tabcolsep}{2pt}
    \begin{tabular}{cc}
      \toprule
      \textbf{Norm} & \textbf{mIoU} \\
      \textbf{Factor} & \\
      \midrule
      1 & 70.41 \\
      $\sqrt{2}$ & 70.96 \\
      $\sqrt{3}$ & 69.59 \\
      $\sqrt{5}$ & 68.95 \\
      \bottomrule
    \end{tabular}
    \vspace{2pt}
    \label{tab:norm-fact}
  \end{minipage}

  \vspace{2pt}
\noindent
\fontsize{8pt}{8pt}\selectfont
\hspace{0.03\textwidth} (a) 
\hspace{0.37\textwidth} (b)
\hspace{0.27\textwidth} (c)

  \vspace{6pt} % space between row 1 and row 2

  %---------------------- Second Row: 2 Tables ----------------------
 \fontsize{6.5pt}{6.5pt}\selectfont
 \begin{minipage}[t]{0.5\textwidth} 
    
  \centering
  \setlength{\tabcolsep}{3pt} 
  \begin{tabular}{c|cccccc}
    \toprule
    \textbf{No. of Exemplars} & 1 & 2 & 3 & 4 & 5 & 6 \\
    \midrule
    \textbf{mIoU} & 65.78 & 69.24 & 70.53 & 70.96 & 70.90 & 70.81 \\
    \bottomrule
  \end{tabular}
  \vspace{2pt}
  \label{dmg}
\end{minipage}%
  \hfill
  \begin{minipage}[t]{0.5\textwidth} 
    \centering
    \setlength{\tabcolsep}{4pt}
    \begin{tabular}{c|cccc}
      \toprule
      \textbf{No. of Iterations} & 1 & 2 & 3 & 4 \\
      \midrule
      \textbf{PerSense (mIoU)} & 70.96 & 70.97 & 70.96 & 70.95 \\  \vspace{2pt}
      \textbf{Avg Inf Time (sec)} & 2.7 & 3.1 & 3.5 & 3.9 \\
      \bottomrule
    \end{tabular}
    \vspace{2pt}
    \label{iterations}
  \end{minipage}

  \vspace{2pt}
\noindent
\fontsize{8pt}{8pt}\selectfont
\hspace{0.01\textwidth} (d)
\hspace{0.45\textwidth} (e)

  \caption{(a) Running efficiency comparison. (b) Component-wise ablation. (c) Normalization factor in PPSM. (d) Number of exemplars in DMG. (e) Multiple feedback iterations.}
  \label{merged-all-tables}
  \vspace{-5pt}
\end{table*}
\vspace{1pt}
 \noindent\textbf{Performance on Sparse Benchmarks:}  
While PerSense targets dense scenes, we also evaluate it on sparse benchmarks COCO-20\textsuperscript{i}~\citep{nguyen2019feature} and LVIS-92\textsuperscript{i}~\citep{gupta2019lvis} (Table~\ref{Comparisonwithcoco}), where DMs are less effective and traditional object detectors perform well due to the simplicity of the task. Despite this, PerSense performs competitively on COCO, surpassing PerSAM-F by +25.5\% and Painter~\citep{wang2023images} by +15.9\%. SegGPT leads on COCO due to training set overlap. On LVIS, PerSense outperforms PerSAM-F by +13.4\%, SegGPT~\citep{wang2023seggpt} by +7.1\%, and Painter by +15.2\%. Overall, it ranks second-best, trailing only Matcher~\citep{liu2023matcher}, demonstrating strong generalization even in scenarios outside its primary design focus.

\noindent\textbf{Model-Agnostic Framework:}  
In addition to demonstrating model-agnostic capability with two different DMGs, PerSense also generalizes to dense scenarios in the medical domain. Fig.~\ref{fig:qualitative_results} presents qualitative results for cellular segmentation leveraging MedSAM~\citep{ma2024segment} encoder and decoder on Modified Bone Marrow (MBM) dataset~\citep{paul2017count}. These results highlight PerSense's broad applicability, spanning domains from industrial to medical.

\noindent\textbf{Runtime Efficiency:}  
Table~\ref{merged-all-tables} reports inference time and memory usage for PerSense on a single NVIDIA RTX 4090 GPU (batch size = 1). PerSense is more efficient than Matcher and PerSAM, and incurs only marginal overhead compared to Grounded-SAM, though with a trade-off in segmentation accuracy. As a model-agnostic framework, PerSense can flexibly integrate lightweight backbones tailored to the target application, allowing users to balance accuracy and efficiency as needed. 

\noindent\textbf{Limitations: } While PerSense employs IDM and PPSM to refine DMs and reject false positives, respectively, 
% and integrates a feedback mechanism to enhance exemplar selection for improved DMG performance, 
it cannot recover any true positives missed initially by DMG, during DM generation. See Appendix (\ref{failure-cases}) for failure cases. 

% Additionally, inference time can be further reduced through optimized implementations.

% Notably, PerSense is model-agnostic and can leverage lightweight, efficient backbones to significantly reduce runtime. Furthermore, its inference time can also be improved through optimized implementations.

% \noindent\textbf{Domain-agnostic Performance:} 

% \noindent\textbf{Runtime Efficiency:}  
% Table~\ref{runningefficiency} reports inference time and memory usage for PerSense on a single NVIDIA RTX 4090 GPU (batch size = 1). PerSense is more efficient than Matcher and PerSAM, and incurs only marginal overhead compared to Grounded-SAM, though with a trade-off in segmentation accuracy.

% \noindent\textbf{Runtime Efficiency:} Table~\ref{runningefficiency} provides inference time and memory consumption details for PerSense, evaluated on a single NVIDIA GeForce RTX 4090 GPU (batch size = 1). PerSense is computationally efficient than Matcher and PerSAM and incurs marginal latency and GPU memory usage compared to Grounded-SAM, though with a trade-off in segmentation accuracy.

% \noindent\textbf{Limitations: } While PerSense employs IDM and PPSM to refine DMs and reject false positives, respectively, 
% % and integrates a feedback mechanism to enhance exemplar selection for improved DMG performance, 
% it cannot recover any true positives missed initially by DMG, during DM generation. See supplementary (S2) for failure cases. 

% \textbf{Runtime Efficiency: } We evaluated the runtime efficiency using single NVIDIA RTX 4090 GPU (batch size = 1). As shown in Table~\ref{tab:memory_inference_comparison}, PerSAM is computationally inefficient due to its iterative masking, requiring one iteration per object instance. Matcher averages 10.2 seconds per image, limiting its applicability for real-world tasks. In contrast, PerSense processes an image in 2.7 seconds. Although The 0.9-second overhead in PerSense is due to DM generation for instance-level point prompts in dense scenes. Overall, PerSense balances efficiency and accuracy, outperforming PerSAM and Matcher while maintaining minimal latency compared to Grounded-SAM.

% \subsection{Ablation Study}
% \label{subsection:ablation_study}
\noindent\textbf{Ablation Study: } Table~\ref{merged-all-tables} presents following ablations:
\textbf{\textit{(a) Component-wise Ablation: }} 
PerSense comprises three core components: IDM, PPSM, and a feedback mechanism. Integrating PPSM with baseline improves mIoU by +1.37\% on PerSense-D and +2.48\% on COCO by reducing false positives in IDM outputs. The feedback mechanism further boosts mIoU by +4.01\% on PerSense-D by refining exemplar selection, but offers limited gain on COCO (+0.19\%) due to its sparsity (limited count). \textbf{\textit{(b) Choice of Normalization Factor in PPSM: }} For PPSM's adaptive threshold, we initialized the normalization constant at 1 and used a square root progression to gradually increase step size. Empirically, $\sqrt{2}$ provided the highest mIoU improvement. \textbf{\textit{(c) Varying Number of Exemplars in Feedback Mechanism:}}  
Using SAM scores, our feedback mechanism automatically selects the best exemplars for DMG. Segmentation performance on PerSense-D peaks with four exemplars, beyond which additional exemplars contribute little to no improvement. \textbf{\textit{(d) Multiple Iterations in Feedback Mechanism:}}  
PerSense refines the DM in a single pass by selecting exemplars from initial segmentation output using SAM scores. Additional iterations yield no accuracy gains but increase computational cost. This occurs because first-pass exemplars, with well-defined boundaries, are already effectively captured by SAM. Subsequent iterations redundantly select the same exemplars due to their distinct features and consistently high SAM scores.

\vspace{-8pt}

\section{Conclusion}
\label{sec:conclusion}
We presented PerSense, a training-free and model-agnostic one-shot framework for personalized instance segmentation in dense images. We proposed IDM and PPSM, which transforms DMs from DMG into personalized instance-level point prompts for segmentation. We also proposed a robust feedback mechanism in PerSense which automates and improves the exemplar selection process in DMG. Finally to promote algorithmic advancements considering the persense task, we introduced PerSense-D, a benchmark exclusive to dense images. We established superiority of our method in dense scenarios by comparing it with the SOTA.

\bibliography{egbib}

\clearpage  % <-- Start Appendix on a new page
\section{Appendix}  % <-- Unnumbered section

\subsection{Algorithms}
\label{algorithm}

\vspace{-10pt}
%PerSense
\begin{algorithm*}[h]
    \small
    \SetAlgoLined
    \KwIn{\textit{Query Image ($I_{Q}$)}, \textit{Support Image ($I_{S}$)}, \textit{Support Mask ($M_{S}$)}}
    \KwOut{\textit{Segmentation Mask}}
    Perform input masking: $I_{\text{masked}} = I_{S} \odot M_{S}$\;
    Extract class-label using CLE from $I_{\text{masked}}$ (text prompt:  "\textit{Name the object in the image?}")\;
    Prompt grounding detector with class-label\;
    Obtain grounded detections\;
    Bounding box with max confidence $\rightarrow$ decoder\;
    Obtain segmentation mask of the object\;
    Refine bounding box coordinates using the segmentation mask\;
    Exemplar Selection: Refined bounding box $\rightarrow$ DMG\;
    Obtain DM from DMG\;
    Process DM using IDM to generate candidate point prompts ($PP_{\text{cand}}$)\;
    $PP_{\text{cand}}$ $\rightarrow$ PPSM $\rightarrow$ final point prompts ($PP_{\text{final}}$)\;
    $PP_{\text{final}}$ $\rightarrow$ decoder\;
    Obtain an initial segmentation output\;
    Select Top 4 candidates as DMG exemplars based on SAM score\;
    Feedback: Repeat Steps 8 to 13\;
    Obtain final segmentation output\;
    \caption{PerSense}
    \label{alg:persense}
\end{algorithm*}
%IDM

%PPSM
% \vspace{-20pt}
\begin{algorithm*}[h]
    \small
    \SetAlgoLined
    \KwIn{candidate\_PP, similarity\_matrix, object\_count, grounded\_detections}
    \KwOut{selected\_PP}
    
    max\_score $\leftarrow$ Get the maximum similarity score from similarity\_matrix\;
    selected\_PP $\leftarrow$ [ ] \tcp*{Empty list to store selected\_PP}
    sim\_threshold $\leftarrow$ max\_score / (object\_count / $\sqrt{2}$)\;
    
    \For{each PP in candidate\_PP}{
        PP\_similarity $\leftarrow$ similarity\_matrix(PP)\;
        \For{each box in grounded\_detections}{
            \If{(PP\_similarity $>$ sim\_threshold) \textbf{and} (PP lies within box)}{
                selected\_PP.append(PP)\;
            }
        }
    }

    \Return selected\_PP\;
    \caption{Point Prompt Selection Module (PPSM)}
    \label{alg:ppsm}
\end{algorithm*}

% \vspace{-60pt}
\begin{algorithm*}[h]
    \SetAlgoLined
    \small
    \KwIn{\textit{Density Map (DM) from DMG}}
    \KwOut{\textit{Candidate Point Prompts (PP)}}
    Convert DM to grayscale image (\textit{I$_\text{gray}$})\;
    Threshold to binary (threshold = 30) to obtain binary image (\textit{I$_\text{binary}$})\;
    Erode \textit{I$_\text{binary}$} using 3 $\times$ 3 kernel\;
    Find BLOB\_contours (\textit{C$_\text{BLOB}$}) in the eroded image (\textit{I$_\text{eroded}$})\;
    \For{contour in \textit{C$_\text{BLOB}$}}{
        Compute contour area (\textit{A$_\text{contour}$})\;
        Find center pixel coordinates for each contour\;
    }
    Compute mean ($\mu$) and standard deviation ($\sigma$) using \textit{A$_\text{contour}$}\;
    Detect composite\_contours (\textit{C$_\text{composite}$}) by thresholding \textit{A$_\text{contour}$}\;
    \quad area\_threshold = $\mu + 2\sigma$\;
    \For{contour in \textit{C$_\text{BLOB}$}}{
        Compute \textit{A$_\text{contour}$}\;
        \If{\textit{A$_\text{contour}$} \textgreater\ area\_threshold}{
            save contour as \textit{C$_\text{composite}$}\;
        }
    }
    \For{contour in \textit{C$_\text{composite}$}}{
        Apply distance transform [threshold = \textit{0.5 * dist\_transform.max()}]\;
        Find child contours\;
        Find center pixel coordinates for each child contour\;
    }
    \Return center points from Steps 7 and 21 as candidate PP\;
    \caption{Instance Detection Module (IDM)}
    \label{alg:idm}
\end{algorithm*}

\flushleft
% \vspace{-30pt}
\subsection{PerSense vs SOTA: Class-wise mIoU Comparison}
\label{classmiou}
\vspace{70pt}

\begin{figure}[h]
    \centering
    \vspace{-80pt}
    \includegraphics[width=0.85\linewidth]{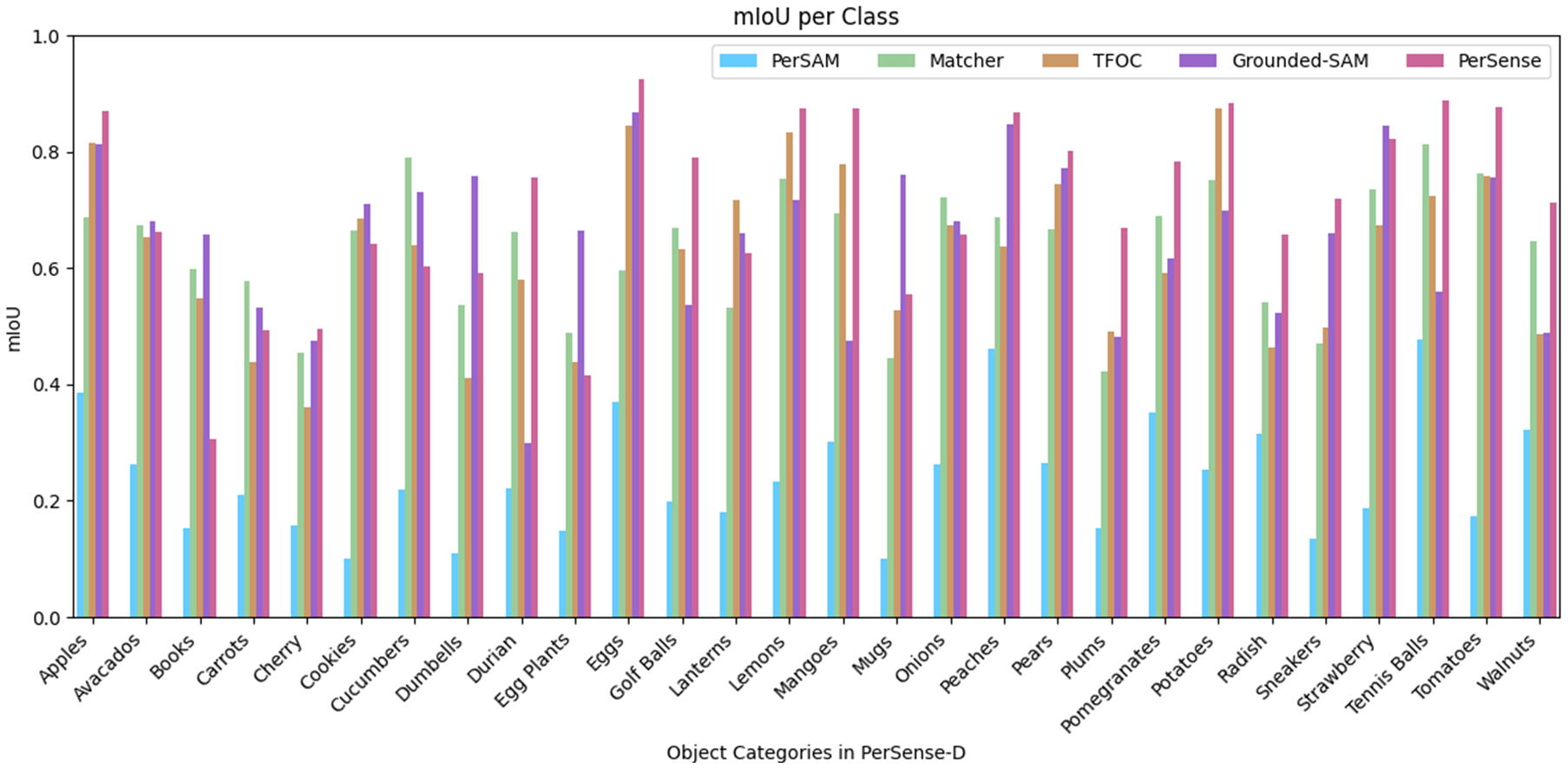}
    \caption{Class-wise mIoU comparison between PerSense and SOTA training-free approaches on the PerSense-D benchmark.}
    \label{fig:classwisemiou}
\end{figure}

\clearpage

\flushleft
\subsection{Failure Cases}
\label{failure-cases}

\begin{figure}[h]
    \centering
    \includegraphics[width=1\linewidth]{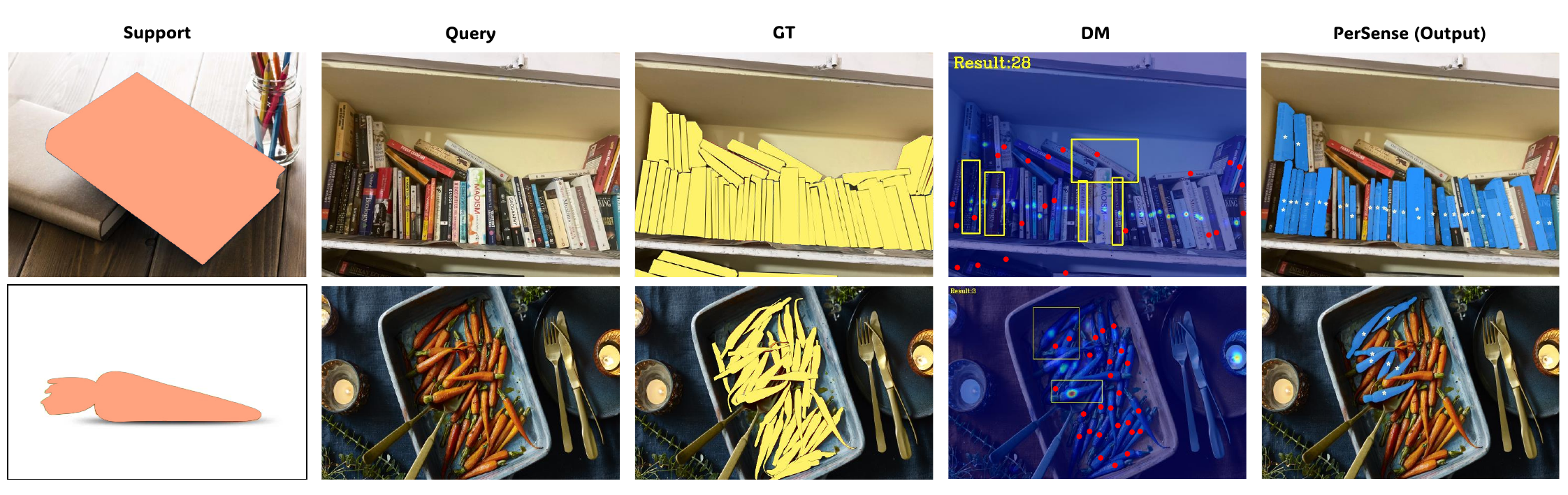}
    \caption{The figure illustrates scenarios where PerSense's performance deteriorates, primarily due to its reliance on the generated density map. In the first row, where the goal is to segment all instances of the "book" class, the density map excludes many true positives (highlighted in red), which PerSense cannot recover once they are lost during DM generation. A similar issue is seen in the second row, where a poor-quality density map for the "carrot" class leads to missed instances, negatively impacting PerSense segmentation performance.}
    \label{fig:failurecases}
\end{figure}

% \newpage
% \clearpage
% \flushleft
\subsection{Annotated Images from PerSense-D}
\label{annotated_images}

\begin{figure}[h]
    \centering
    \vspace{-10pt}
    \includegraphics[width=1\linewidth]{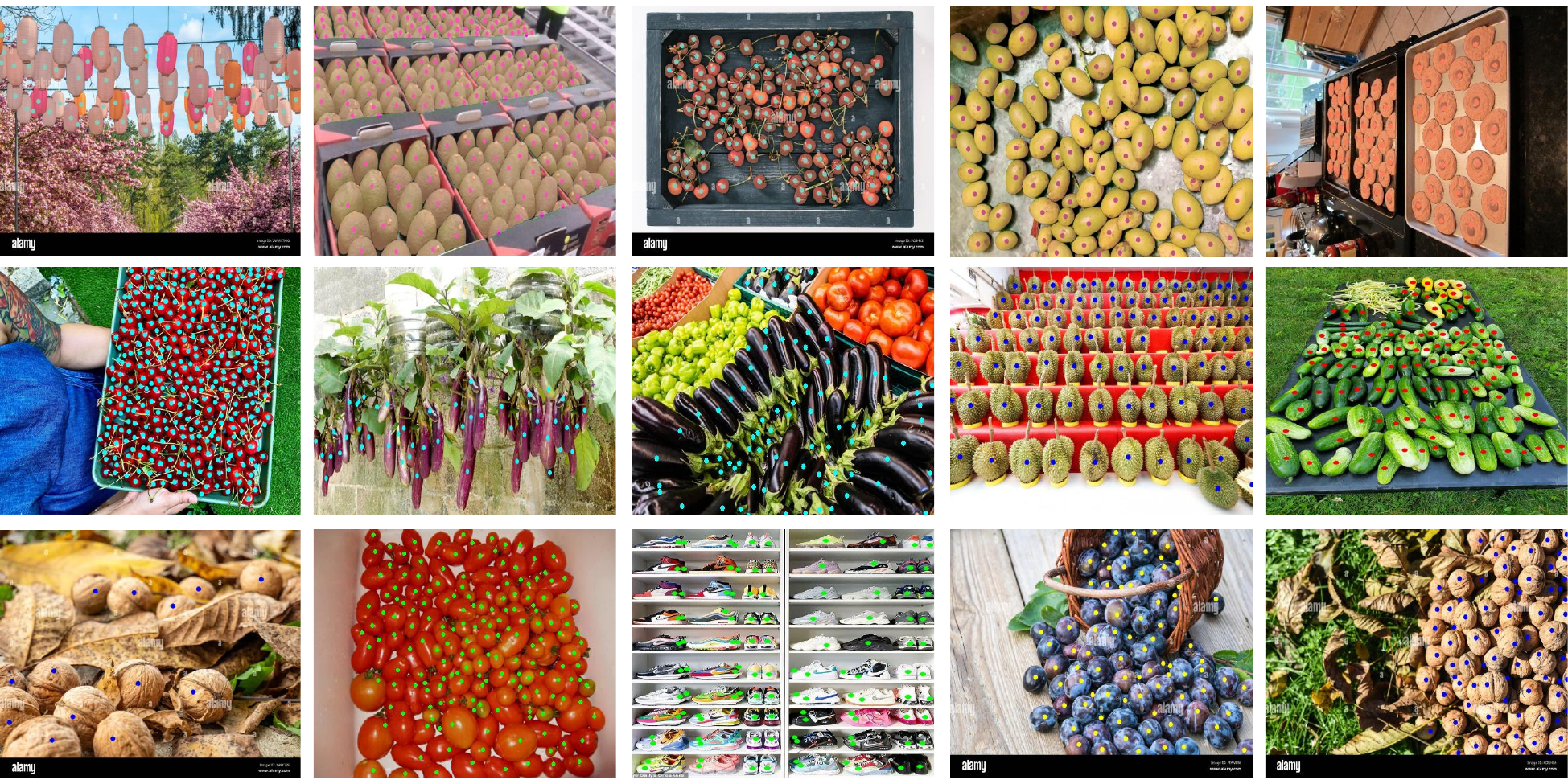}
    \caption{Few annotated images from PerSense-D. Each image presents dot and mask annotations, respectively.}
\end{figure}

\end{document}